\documentstyle[twoside,wicsbook]{article}

\newcommand{\type}{ \,{\bf :}\, }
\newcommand{\memberof}{\,{\in}\,}
\newcommand{\tensorimplysource}
   {\! \mbox{\hspace{.21em}--\hspace{-.21em}} \backslash }

\newcommand{\interior}[2]{{#1}^{\prime\prime}_{#2}}

\newcommand{\define}{\stackrel{{\rm df}}{=}}
\newcommand{\product}[2]{ {#1} {\times} {#2} }
\newcommand{\derivedir}[2]{ {#1}^{\prime}_{#2} }
\newcommand{\deriveinv}[2]{ {#1}^{\prime}_{#2} }
\newcommand{\pair}[2]{\langle #1,#2 \rangle}
\newcommand{\triple}[3]{\mbox{$ \langle #1,#2,#3 \rangle $}}
\newcommand{\quadruple}[4]{\mbox{$ \langle #1,#2,#3,#4 \rangle $}}

\title{\vspace{-3pc}\titlesize\bf Rough Concept Analysis}
 
\author{
	\large Robert E.\ Kent \\
	\normalsize University of Arkansas at Little Rock \\
	\normalsize Little Rock, Arkansas, U.S.A.
\date{}}

\begin{document}

\maketitle

\begin{abstract}\ninesize
\noindent The theory introduced, presented and developed in this paper,
is concerned with Rough Concept Analysis.
This theory is a synthesis 
of   the theory of Rough Sets pioneered by Zdzislaw Pawlak \cite{pawlak82}
with the theory of Formal Concept Analysis pioneered by Rudolf Wille \cite{wille82}.
The central notion in this paper of a {\em rough formal concept\/}
combines in a natural fashion
the notion of a rough set with the notion of a formal concept:
	``rough set $+$ formal concept $\;=\;$ rough formal concept''.
A related paper \cite{kent94c} provides a synthesis of the two important data modeling techniques:
conceptual scaling of Formal Concept Analysis,
and Entity-Relationship database modeling.
\end{abstract}

\section{The RS-FCA Community}
\label{thers-fcacommunity}

The theory of Rough Sets initiated by Zdzislaw Pawlak \cite{pawlak82}
is used to model imprecise or incomplete knowledge and approximate classification.
The theory of Formal Concept Analysis initiated by Rudolf Wille \cite{wille82}
is used for data modeling, analysis and interpretation,
and also for knowledge representation and knowledge discovery
via the special technique of attribute exploration
or the more general technique of concept exploration.
Rough Sets and Formal Concept Analysis have much in common,
both in terms of goals and methodologies.
Various analogies ($\cong$) and identities ($\equiv$) 
between Rough Sets notions and Formal Concept Analysis notions
are listed here.
\begin{center}
\begin{tabular}{|c@{\hspace{2mm}}c@{\hspace{2mm}}c|} \hline
	{\bf Rough Sets} 
	 &
	  & {\bf Formal Concept Analysis} 
	   \\ \hline\hline
	approximation space \cite{pawlak82}
	 & $\equiv$ 
	  & formal context morphism 
	   \\ \hline
	indexed collection of subsets 
	 & $\equiv$ 
	  & formal context \cite{wille82}
	   \\ \hline
	--- 
	 & $\equiv$ 
	  & concept lattice 
	   \\ \hline
	information system 
	 & $\equiv$ 
	  & many-valued formal context \cite{kent94c}
	   \\ \hline
	$\left. \mbox{\begin{tabular}{c} discretization 
	                                 \\ attribute-value pairing 
	                                 \\ subranging \\ complementation 
	                                 \\ $\vdots$ 
	              \end{tabular}} \right\}$
	 & $\cong$ 
	  & \begin{tabular}{c} interpretation via constraints \cite{kent93b,kent94c}
	                       \\ (special case: conceptual scaling) 
	    \end{tabular}
	   \\ \hline
	certain rule
	 & $\equiv$ 
	  & implication 
	   \\ \hline
	possible rule
	 & $\equiv$ 
	  & certain rule with dichotomic scaling
	   \\ \hline
	rough measure of rule
	 & $\cong$ 
	  & $\left\{ \mbox{\begin{tabular}{c} rough measure of implication \\ $U \stackrel{k}{\Rightarrow} V$ when $\frac{|\deriveinv{U}{} \cap \deriveinv{V}{}|}{|\deriveinv{U}{}|} = k$ \\ partial implication \end{tabular}} \right.$
	   \\ \hline
\end{tabular}
\end{center}
Rough Sets works directly with
information systems $\equiv$ many-valued formal contexts,
and only implicitly with a derived structure containing attribute-value pairs,
whereas Formal Concept Analysis has an explicit transformation
[many-valued formal context $\Rightarrow$ (ordinary) formal context]
called conceptual scaling,
which is regarded as an act of interpretation.

\section{Rough Formal Contexts}
\label{roughformalcontexts}

This paper introduces the new theory of Rough Concept Analysis,
which is a synthesis of Rough Sets and Formal Concept Analysis.

An {\em approximation space\/} is a pair $\pair{G}{{E}}$,
where $G$ is a set of {\em objects\/} or {\em entities\/}
and ${E}$ is an equivalence relation on $G$ called an {\em indiscernibility relation\/}.
A {\em formal context\/} is a triple $\triple{G}{M}{I}$
consisting of 
a set of {\em objects\/} $G$,
a set of {\em attributes\/} $M$,
and a binary {\em incidence\/} relation $I \subseteq \product{G}{M}$ between $G$ and $M$,
where $g{I}m$ asserts that ``object $g$ {\em has\/} attribute $m$''
for any object $g \memberof G$ and attribute $m \memberof M$.
A {\em formal concept\/} of a given formal context 
will consist of an extent/intent pair
$({A},{B})$
where
the intent 
$B = \derivedir{A}{I}
 \define \{ m \memberof M \mid g{I}m \mbox{ for all } g \memberof A \}
 = \bigcap_{g \in A} gI 
 \subseteq M$ 
contains precisely 
those attributes shared by all objects in the extent $A$,
and vice-versa,
the extent 
$A = \deriveinv{B}{I}
 \define \{ g \memberof G \mid g{I}m \mbox{ for all } m \memberof B \}
 = \bigcap_{m \in B} Im
 \subseteq G$ 
contains precisely 
those objects sharing all attributes in the intent $B$.
The collection of all concepts is ordered by generalization-specialization.
Concepts with the generalization-specialization ordering
form a complete lattice ${\cal B}\triple{G}{M}{I}$
called the {\em concept lattice\/} of $\triple{G}{M}{I}$.

Given any approximation space $\pair{G}{E}$ on objects
and given any formal context $\triple{G}{M}{I}$,
an attribute $m \memberof M$ is a {\em definable attribute\/}
when
its extent $Im \subseteq G$ is a definable subset of objects
w.r.t.\ indiscernibility relation $E$.
A {\em definable formal context\/} is a context
all of whose attributes are definable.
For any formal context $\triple{G}{M}{I}$
we wish to approximate $I$ in terms of definable contexts.
We use two notions for this:
an upper approximation of possibility and a lower approximation of necessity.
These two {\em contextual\/} approximations provide
upper and lower {\em conceptual\/} approximations 
for concepts in ${\cal B}\triple{G}{M}{I}$
(see Section~\ref{roughformalconcepts}).

Let $\pair{G}{E}$ be a fixed approximation space on objects $G$.
\begin{description}
	\item[{[Upper $E$-approximation]}]
		The upper $E$-approximation of $I$,
		denoted by $\overline{I}^E$,
		is defined element-wise:
		for each attribute $m \memberof M$,
		the extent of $m$ in the upper approximation $\overline{I}^E$
		is the upper approximation of its extent in $I$,
		${\overline{I}^E}m
		 \define \overline{Im}^E
		 = \{ g \mid [g]_E \cap Im \not= \emptyset \}$.
		This upper approximation is the left relational composition of $I$ by $E$
		$\overline{I}^E = E \circ I = [\;] \circ \exists_{[\;]}(I)$.
		The upper approximation of $I$ is the least definable context containing $I$.
		The extent of a subset of attributes $B \subseteq M$ with respect to the upper $E$-approximation is
		$\deriveinv{B}{\overline{I}^E} 
		 = \bigcap_{m \in B} {\overline{I}^E}m
		 = \bigcap_{m \in B} \overline{Im}^E 
		 \supseteq \overline{\bigcap_{m \in B} Im}^E
		 = \overline{\deriveinv{B}{I}}^E$
		(see the important discussion below about choices).
		The upper $E$-approximation is a monotonic function
		$\overline{(\;)}^E \type \mbox{\bf Cxt}_{G} \rightarrow \mbox{\bf Cxt}_{G}$.
	\item[{[Lower $E$-approximation]}]
		The lower $E$-approximation of $I$,
		denoted by $\underline{I}_E$,
		is also defined element-wise:
		for each attribute $m \memberof M$,
		the extent of $m$ in the lower approximation $\underline{I}_E$
		is the lower approximation of its extent in $I$,
		${\underline{I}_E}m
		 \define \underline{Im}_E
		 = \{ g \mid [g]_E \subseteq Im \}$.
		This lower approximation is the left relational residuation of $I$ by $E$
		$\underline{I}_E = E \tensorimplysource I = [\;] \circ \forall_{[\;]}(I)$.
		The lower approximation of $I$ is the greatest definable context contained in $I$.
		The extent of a subset of attributes $B \subseteq M$ with respect to the lower $E$-approximation is
		$\deriveinv{B}{\underline{I}_E}
		 = \bigcap_{m \in B} {\underline{I}_E}m
		 = \bigcap_{m \in B} \underline{Im}_E
		 = \underline{\bigcap_{m \in B} Im}_E
		 = \underline{\deriveinv{B}{I}}_E$.
		Lower $E$-approximation is a monotonic function
		$\underline{(\;)}_E \type \mbox{\bf Cxt}_{G} \rightarrow \mbox{\bf Cxt}_{G}$.
\end{description}
There will be some controversy concerning
which definition is better
for the upper approximation of a collection of attributes.
The choices are as follows.
\begin{enumerate}
	\item The stricter choice
		$\overline{\deriveinv{B}{I}}^E
		 = \overline{\bigcap_{m \in B} Im}^E
		 = \bigcup \{ [g] \mid [g] \cap \left(\bigcap_{m \in B} Im\right) \not= \emptyset \}$
		includes only those equivalence classes
		which contain an element of the extent intersection.
	\item The freer choice
		$\deriveinv{B}{\overline{I}^E}
		 = \bigcap_{m \in B} \overline{Im}^E
		 = \bigcap_{m \in B} \left( \bigcup \{ [g] \mid [g] \cap Im \not= \emptyset \} \right)$
		 additionally includes those equivalence classes
		 which contain elements in each individual extent,
		 but do not contain an element in the combined extent intersection.
		 In this paper we have chosen this freer definition,
		 partly for better mathematical tractibility,
		 partly because it corresponds to direct existential image of contexts from formal concept analysis,
		 and
		 partly because it by itself has a valid semantics.
\end{enumerate}

For any object $g \memberof G$ and any subset $B \subseteq M$,
we say that
$g$ {\em certainly has\/} all attributes in $B$
when
$g \memberof \deriveinv{B}{\underline{I}_E} = \underline{\deriveinv{B}{I}}_E$,
and that
$g$ {\em possibly has\/} all attributes in $B$
when
$g \memberof \deriveinv{B}{\overline{I}^E} \supseteq \overline{\deriveinv{B}{I}}^E$.
There are three ordering relations for contexts:
the {\em upper\/} (Smyth) {\em order\/}  $I \leq^u J$ iff $\overline{I}^E \subseteq \overline{J}^E$,
the {\em lower\/} (Hoare) {\em order\/}  $I \leq^l J$ iff $\underline{I}_E \subseteq \underline{J}_E$,
and
the {\em rough\/} (Milner) {\em order\/} $I \leq J$ iff $I \leq^l J$  and $I \leq^u J$
iff $\underline{I}_E \subseteq \underline{J}_E$ and $\overline{I}^E \subseteq \overline{J}^E$.
Two contexts $\triple{G}{M}{I}$ and $\triple{G}{M}{J}$ of $G$-objects and $M$-attributes
are {\em $E$-roughly equal\/},
denoted by $I \equiv J$,
when both $I \leq J$ and $J \leq I$.
The rough order $\leq$ is only a preorder:
it is reflexive and transitive,
but it is not necessarily antisymmetric.
To make the rough order into a partial order 
and to change rough equality into true equality,
we must ``quotient out'' by rough equality.
A {\em rough formal context\/} in $\pair{G}{E}$
is
a collection of roughly equal formal contexts of $G$-objects and $M$-attributes;
or,
equivalently,
a rough context is a collection of formal contexts of $G$-objects and $M$-attributes
which have the same upper and lower approximation contexts.
Any quadruple $\quadruple{G}{E}{M}{I}$,
consisting of a formal context $\triple{G}{M}{I}$ and an approximation space $\pair{G}{E}$ on its set of objects,
can be regarded as the rough formal context 
consisting of all contexts roughly equal to $I$.

\section{Rough Formal Concepts}
\label{roughformalconcepts}

In this section we define the notions of ``approximation'' and ``rough equality'' with respect to formal concepts.
Given any approximation space $\pair{G}{E}$ on objects
and given any formal context $\triple{G}{M}{I}$,
a formal concept $(A,B) \in {\cal B}\triple{G}{M}{I}$ is a {\em definable concept\/}
when
its extent $A \subseteq G$ is a definable subset of objects
w.r.t.\ indiscernibility relation $E$.
All concepts of a definable formal context are definable formal concepts.
Let $\triple{G}{M}{I}$ be any formal context with an approximation space $\pair{G}{E}$ on objects.
We wish to approximate concepts in ${\cal B}\triple{G}{M}{I}$ in terms of $E$-definable concepts.
We do this {\em externally\/} in terms of concepts of the upper and lower approximation contexts of $I$
--- two $E$-definable formal contexts.
Just as for subsets and contexts,
we use two notions for approximating concepts.
\begin{description}
	\item[{[Upper $E$-approximation]}]
		The {\em upper $E$-approximation\/} of a concept $(A,B) \in {\cal B}\triple{G}{M}{I}$
		is the concept $\overline{(A,B)}^E \in {\cal B}\triple{G}{M}{\overline{I}^E}$
		defined by
		$\overline{(A,B)}^E 
		 = (\deriveinv{B}{\overline{I}^E},\interior{B}{\overline{I}^E})$.
		Upper $E$-approximation is a monotonic function
		$\overline{(\;)}^E$
		which assigns 
		concepts in the upper approximation concept lattice 
		${\cal B}\triple{G}{M}{\overline{I}^E}$
		to 
		concepts in 
		${\cal B}\triple{G}{M}{I}$.
	\item[{[Lower $E$-approximation]}]
		The {\em lower $E$-approximation\/} of a concept $(A,B) \in {\cal B}\triple{G}{M}{I}$
		is the concept $\underline{(A,B)}_E \in {\cal B}\triple{G}{M}{\underline{I}_E}$
		defined by
		$\underline{(A,B)}_E 
		 = (\deriveinv{B}{\underline{I}_E},\interior{B}{\underline{I}_E})$.
		Lower $E$-approximation is a monotonic function
		$\underline{(\;)}_E$
		which assigns 
		concepts in the lower approximation concept lattice 
		${\cal B}\triple{G}{M}{\underline{I}_E}$
		to 
		concepts in 
		${\cal B}\triple{G}{M}{I}$.
\end{description}
Upper approximation assignment
is left adjoint to
a lower-join operator
$\overline{\left( \rule{0cm}{2ex}\; \right)}^E \dashv {\bigvee}_E$
defined by
$\bigvee_E (A,B) \define \bigvee \{ (A_1,B_1) \in {\cal B}\triple{G}{M}{I} \mid A_1 \subseteq A \}$
for all concepts $(A,B) \in {\cal B}\triple{G}{M}{\overline{I}^E}$.
Hence,
upper approximation assignment
is a join-preserving monotonic function.
Lower approximation assignment
is right adjoint to
an upper-meet operator
${\bigwedge}_E \dashv \underline{\left( \rule{0cm}{2ex}\; \right)}_E$
defined by
$\bigwedge_E (A,B) \define \bigwedge \{ (A_1,B_1) \in {\cal B}\triple{G}{M}{I} \mid A \subseteq A_1 \}$
for all concepts $(A,B) \in {\cal B}\triple{G}{M}{\underline{I}^E}$.
Hence,
lower approximation assignment
is a meet-preserving monotonic function.

Just as for subsets and contexts,
there are three ordering relations for concepts:
for any two formal concepts $C_1 = (A_1,B_1)$ and $C_2 = (A_2,B_2)$,
the {\em upper\/} (Smyth) {\em order\/}  $C_1 \leq^u C_2$ iff $\overline{C_1}^E \subseteq \overline{C_2}^E$,
the {\em lower\/} (Hoare) {\em order\/}  $C_1 \leq^l C_2$ iff $\underline{C_1}_E \subseteq \underline{C_2}_E$,
and
the {\em rough\/} (Milner) {\em order\/} $C_1 \leq C_2$ iff $C_1 \leq^l C_2 \mbox{ and } C_1 \leq^u C_2$
iff $\underline{C_1}_E \subseteq \underline{C_2}_E$ and $\overline{C_1}^E \subseteq \overline{C_2}^E$.
Two concepts $C_1$ and $C_2$ of a formal context $\triple{G}{M}{I}$
are {\em $E$-roughly equal\/},
denoted by $C_1 \equiv C_2$,
when both $C_1 \leq C_2$ and $C_2 \leq C_1$.
Again the rough order $\leq$ is only a preorder.
To make the rough order into a partial order and to change rough equality into true equality,
we must ``quotient out'' by rough equality.
A {\em rough concept\/} of a formal context $\triple{G}{M}{I}$ with approximation space $\pair{G}{E}$
is
a collection of roughly equal concepts;
or,
equivalently,
a rough concept is a collection of concepts
which have the same upper and lower conceptual approximations.
Rough concepts for $E$-definable contexts are crisp,
since two concepts are roughly equal iff they are precisely equal.

\section{Example}
\label{example}

In Table~\ref{Living} is an example of a formal context called the Living context.
This formal context is concerned with
a simple ecological description of some living organisms.
Although somewhat simplistic,
it is quite useful for illustrative purposes.
This was one of several formal contexts presented in a seminar by Rudolf Wille
at the University of Arkansas in 1992.
It was originally taken from an Hungarian children's television show.
We provide a rough conceptual analysis of the Living context in this paper.
Table~\ref{Living} also contains the concept lattice for the Living context.
The 19 formal concepts of the Living context
$\{ \mbox{B}_0,\mbox{B}_1,\ldots,\mbox{B}_{18} \}$
represented by indices in Figure~\ref{Living},
include the top formal concept $\mbox{B}_0$ representing ``all Living organisms'',
the bottom formal concept $\mbox{B}_{18}$ with ``no Living organisms'',
and formal concepts such as $\mbox{B}_6$ representing ``limbed land organisms'',
whose intent consists of the attributes
``needs water'',
``is motile'',
``has limbs'' and 
``lives on land'',
and whose extent consists of the organisms
``Dog'' and 
``Frog''.
\scriptsize
\begin{table}
\begin{center}
\begin{tabular}{c@{\hspace{5mm}}c}
\begin{tabular}{c}
	\begin{tabular}[t]{|ll|} \hline
		\multicolumn{2}{|c|}{{\bf object set}} \\ \hline\hline
		Le & Leech \\
		Br & Bream \\
		Fr & Frog  \\
		Dg & Dog   \\
		SW & Spike-Weed \\
		Rd & Reed  \\
		Bn & Bean  \\
		Mz & Maize \\ \hline
	\end{tabular}
\\
	\begin{tabular}[t]{|ll|} \hline
		\multicolumn{2}{|c|}{{\bf attribute set}} \\ \hline\hline
		nw & needs water        \\
		lw & lives in water     \\
		ll & lives on land      \\
		nc & needs chlorophyll  \\
		2lg & 2 leaf germination \\
		1lg & 1 leaf germination \\
		mo & is motile          \\
		lb & has limbs          \\
		sk & suckles young      \\ \hline
	\end{tabular}
\end{tabular}
&
\begin{tabular}{c}
	\begin{tabular}[b]{|l|c@{\hspace{3pt}}c@{\hspace{3pt}}c@{\hspace{3pt}}c@{\hspace{3pt}}
			      c@{\hspace{3pt}}c@{\hspace{3pt}}c@{\hspace{3pt}}c@{\hspace{3pt}}c|} \hline
		\multicolumn{10}{|c|}{{\bf incidence relation}} \\ \hline\hline
		   & nw & lw & ll & nc & 2lg & 1lg & mo & lb & sk \\ \hline
		Le &$\times$&$\times$&&&&&$\times$&& \\
		Br &$\times$&$\times$&&&&&$\times$&$\times$& \\
		Fr &$\times$&$\times$&$\times$&&&&$\times$&$\times$& \\
		Dg &$\times$&&$\times$&&&&$\times$&$\times$&$\times$ \\
		SW &$\times$&$\times$&&$\times$&&$\times$&&& \\
		Rd &$\times$&$\times$&$\times$&$\times$&&$\times$&&& \\
		Bn &$\times$&&$\times$&$\times$&$\times$&&&& \\
		Mz &$\times$&&$\times$&$\times$&&$\times$&&& \\ \hline
	\end{tabular}
\\ \\
	\begin{tabular}[b]{|c|} \hline
		{\bf concept lattice}
		\\ \hline\hline
\setlength{\unitlength}{1pt}
\newcommand{\puttext}[3]{\put(#1,#2){{\mbox{\tiny$#3$\normalsize}}}}
\newcommand{\putdisk}[3]{\put(#1,#2){\circle*{#3}}}
\begin{picture}(190,180)(3,-7)
	\puttext{90}{160}{{\bf }}		
	\putdisk{100}{160}{7}			
	\puttext{90}{160}{{\bf 0}}		
	\puttext{105}{165}{{\rm nw}}		
	\putdisk{40}{120}{7}			
	\puttext{30}{120}{{\bf 1}}		
	\put(40,120){\line(3,2){60}}		
	\puttext{45}{130}{{\rm mo}}		
	\putdisk{80}{120}{7}			
	\puttext{70}{120}{{\bf 5}}		
	\put(80,120){\line(1,2){20}}		
	\puttext{85}{125}{{\rm ll}}		
	\putdisk{120}{120}{7}			
	\puttext{105}{120}{{\bf 11}}		
	\put(120,120){\line(-1,2){20}}		
	\puttext{125}{125}{{\rm lw}}		
	\putdisk{160}{120}{7}			
	\puttext{150}{120}{{\bf 3}}		
	\put(160,120){\line(-3,2){60}}		
	\puttext{165}{125}{{\rm nc}}		
	\putdisk{30}{100}{7}			
	\puttext{20}{100}{{\bf 2}}		
	\put(30,100){\line(1,2){10}}		
	\puttext{35}{105}{{\rm lb}}		
	\putdisk{170}{100}{7}			
	\puttext{160}{100}{{\bf 4}}		
	\put(170,100){\line(-1,2){10}}		
	\puttext{175}{105}{{\rm 1lg}}		
	\putdisk{20}{80}{7}			
	\puttext{10}{80}{{\bf 6}}		
	\put(20,80){\line(1,2){10}}		
	\put(20,80){\line(3,2){60}}		
	\putdisk{60}{80}{7}			
	\puttext{45}{80}{{\bf 12}}		
	\put(60,80){\line(-1,2){20}}		
	\put(60,80){\line(3,2){60}}		
	\puttext{65}{75}{{\rm Le}}		
	\putdisk{100}{80}{7}			
	\puttext{85}{80}{{\bf 15}}		
	\put(100,80){\line(-1,2){20}}		
	\put(100,80){\line(1,2){20}}		
	\putdisk{140}{80}{7}			
	\puttext{130}{80}{{\bf 8}}		
	\put(140,80){\line(-3,2){60}}		
	\put(140,80){\line(1,2){20}}		
	\putdisk{180}{80}{7}			
	\puttext{165}{80}{{\bf 14}}		
	\put(180,80){\line(-3,2){60}}		
	\put(180,80){\line(-1,2){10}}		
	\puttext{185}{75}{{\rm SW}}		
	\putdisk{50}{60}{7}			
	\puttext{35}{60}{{\bf 13}}		
	\put(50,60){\line(-1,2){20}}		
	\put(50,60){\line(1,2){10}}		
	\puttext{55}{55}{{\rm Br}}		
	\putdisk{150}{60}{7}			
	\puttext{140}{60}{{\bf 9}}		
	\put(150,60){\line(-1,2){10}}		
	\put(150,60){\line(1,2){20}}		
	\puttext{155}{55}{{\rm Ma}}		
	\putdisk{0}{40}{7}			
	\puttext{-10}{40}{{\bf 7}}		
	\put(0,40){\line(1,2){20}}		
	\puttext{5}{45}{{\rm sk}}		
	\puttext{5}{30}{{\rm Dg}}		
	\putdisk{40}{40}{7}			
	\puttext{25}{40}{{\bf 16}}		
	\put(40,40){\line(-1,2){20}}		
	\put(40,40){\line(1,2){10}}		
	\put(40,40){\line(3,2){60}}		
	\puttext{45}{38}{{\rm Fr}}		
	\putdisk{120}{40}{7}			
	\puttext{105}{40}{{\bf 10}}		
	\put(120,40){\line(1,2){20}}		
	\puttext{125}{45}{{\rm 2lg}}		
	\puttext{125}{35}{{\rm Bn}}		
	\putdisk{160}{40}{7}			
	\puttext{145}{40}{{\bf 17}}		
	\put(160,40){\line(-3,2){60}}		
	\put(160,40){\line(-1,2){10}}		
	\put(160,40){\line(1,2){20}}		
	\puttext{165}{35}{{\rm Rd}}		
	\putdisk{100}{0}{7}			
	\puttext{85}{-5}{{\bf 18}}		
	\put(100,0){\line(-5,2){100}}		
	\put(100,0){\line(-3,2){60}}		
	\put(100,0){\line(1,2){20}}		
	\put(100,0){\line(3,2){60}}		
\end{picture}
	\\ \hline
	\end{tabular}
\end{tabular}
\end{tabular}
\end{center}
\caption{{\bf the Living formal context and its concept lattice} \label{Living}}
\end{table}
\normalsize

\begin{equation}
	\{
	\{ \mbox{Leech}, \mbox{Bream}, \mbox{Frog} \},
	\{ \mbox{Dog} \},
	\{ \mbox{Spike-Weed}, \mbox{Reed} \},
	\{ \mbox{Bean}, \mbox{Maize} \}
	\}
	\label{indiscernibility}
\end{equation}
is an example of an indiscernibility relation
which forms an approximation space on the objects of the Living context
of Table~\ref{Living}.
This indiscernibility relation is determined by the two conditions:
``lives in water'' and  ``needs chlorophyll''
--- for example,
$\{ \mbox{Bean}, \mbox{Maize} \}$ are those Living organisms
which do not ``live in water'' but do ``need chlorophyll''.
The upper and lower approximation contexts with respect to this indiscernibility relation,
are displayed in Table~\ref{Living:approximation}.
Note that the attributes
``needs water'', ``lives in water'', ``needs chlorophyll'',
``motile'', and ``suckles young''
are all definable attributes.
Clearly,
mutually indiscernible objects are equivalent in the two approximation tables in Table~\ref{Living:approximation},
and can be replaced by their equivalence class,
$[\mbox{Fr}] = \{ \mbox{Leech}, \mbox{Bream}, \mbox{Frog} \}$,
$[\mbox{Dg}] = \{ \mbox{Dog} \}$,
$[\mbox{Rd}] = \{ \mbox{Spike-Weed}, \mbox{Reed} \}$,
and
$[\mbox{Bn}] = \{ \mbox{Bean}, \mbox{Maize} \}$.
The concept lattices for the two approximation contexts of Table~\ref{Living:approximation}
are also displayed there.
\scriptsize
\begin{table}
\begin{center}
\begin{tabular}{c@{\hspace{1cm}}c}
\begin{tabular}{c}
	\begin{tabular}[b]{|l|c@{\hspace{3pt}}c@{\hspace{3pt}}c@{\hspace{3pt}}c@{\hspace{3pt}}
			      c@{\hspace{3pt}}c@{\hspace{3pt}}c@{\hspace{3pt}}c@{\hspace{3pt}}c|} \hline
		\multicolumn{10}{|c|}{{\bf upper approximation}} \\ \hline
		\multicolumn{10}{c}{} \\ \hline
		$\overline{I}^E$ & nw & lw & ll & nc & 2lg & 1lg & mo & lb & sk \\ \hline\hline
		Le &$\times$&$\times$&$\times$&&&&$\times$&$\times$& \\
		Br &$\times$&$\times$&$\times$&&&&$\times$&$\times$& \\
		Fr &$\times$&$\times$&$\times$&&&&$\times$&$\times$& \\ \hline
		Dg &$\times$&&$\times$&&&&$\times$&$\times$&$\times$ \\ \hline
		SW &$\times$&$\times$&$\times$&$\times$&&$\times$&&& \\
		Rd &$\times$&$\times$&$\times$&$\times$&&$\times$&&& \\ \hline
		Bn &$\times$&&$\times$&$\times$&$\times$&$\times$&&& \\
		Mz &$\times$&&$\times$&$\times$&$\times$&$\times$&&& \\ \hline
	\end{tabular}
\\ \\
\setlength{\unitlength}{0.7pt}
\newcommand{\puttext}[3]{\put(#1,#2){{\mbox{\tiny$#3$\normalsize}}}}
\newcommand{\putdisk}[3]{\put(#1,#2){\circle*{#3}}}
\begin{picture}(180,130)(0,0)
	\puttext{70}{120}{{\bf }}		
	\putdisk{80}{120}{7}			
	\puttext{70}{120}{{\bf 0}}		
	\puttext{85}{125}{{\rm nw}}		
	\puttext{85}{131}{{\rm ll}}		
	\putdisk{20}{80}{7}			
	\puttext{10}{80}{{\bf 1}}		
	\put(20,80){\line(3,2){60}}		
	\puttext{25}{88}{{\rm mo}}		
	\puttext{25}{94}{{\rm lb}}		
	\putdisk{100}{80}{7}			
	\puttext{90}{80}{{\bf 5}}		
	\put(100,80){\line(-1,2){20}}		
	\puttext{105}{85}{{\rm lw}}		
	\putdisk{140}{80}{7}			
	\puttext{130}{80}{{\bf 3}}		
	\put(140,80){\line(-3,2){60}}		
	\puttext{145}{85}{{\rm nc}}		
	\puttext{145}{91}{{\rm 1lg}}		
	\putdisk{0}{40}{7}			
	\puttext{-10}{40}{{\bf 2}}		
	\put(0,40){\line(1,2){20}}		
	\puttext{5}{45}{{\rm sk}}		
	\puttext{5}{27}{{\rm Dg}}		
	\putdisk{40}{40}{7}			
	\puttext{30}{40}{{\bf 6}}		
	\put(40,40){\line(-1,2){20}}		
	\put(40,40){\line(3,2){60}}		
	\puttext{45}{35}{{\rm Le}}		
	\puttext{45}{28}{{\rm Br}}		
	\puttext{45}{21}{{\rm Fr}}		
	\putdisk{120}{40}{7}			
	\puttext{110}{40}{{\bf 4}}		
	\put(120,40){\line(1,2){20}}		
	\puttext{125}{45}{{\rm 2lg}}		
	\puttext{125}{35}{{\rm Bn}}		
	\puttext{125}{28}{{\rm Mz}}		
	\putdisk{160}{40}{7}			
	\puttext{150}{40}{{\bf 7}}		
	\put(160,40){\line(-3,2){60}}		
	\put(160,40){\line(-1,2){20}}		
	\puttext{165}{35}{{\rm SW}}		
	\puttext{165}{28}{{\rm Rd}}		
	\putdisk{80}{0}{7}			
	\puttext{70}{-5}{{\bf 8}}		
	\put(80,0){\line(-2,1){80}}		
	\put(80,0){\line(-1,1){40}}		
	\put(80,0){\line(1,1){40}}		
	\put(80,0){\line(2,1){80}}		
\end{picture}
\end{tabular}
&
\begin{tabular}{c}
	\begin{tabular}[b]{|l|c@{\hspace{3pt}}c@{\hspace{3pt}}c@{\hspace{3pt}}c@{\hspace{3pt}}
			      c@{\hspace{3pt}}c@{\hspace{3pt}}c@{\hspace{3pt}}c@{\hspace{3pt}}c|} \hline
		\multicolumn{10}{|c|}{{\bf lower approximation}} \\ \hline
		\multicolumn{10}{c}{} \\ \hline
		$\underline{I}_E$ & nw & lw & ll & nc & 2lg & 1lg & mo & lb & sk \\ \hline\hline
		Le &$\times$&$\times$&&&&&$\times$&& \\
		Br &$\times$&$\times$&&&&&$\times$&& \\
		Fr &$\times$&$\times$&&&&&$\times$&& \\ \hline
		Dg &$\times$&&$\times$&&&&$\times$&$\times$&$\times$ \\ \hline
		SW &$\times$&$\times$&&$\times$&&$\times$&&& \\
		Rd &$\times$&$\times$&&$\times$&&$\times$&&& \\ \hline
		Bn &$\times$&&$\times$&$\times$&&&&& \\
		Mz &$\times$&&$\times$&$\times$&&&&& \\ \hline
	\end{tabular}
\\ \\
\setlength{\unitlength}{0.7pt}
\newcommand{\puttext}[3]{\put(#1,#2){{\mbox{\tiny$#3$\normalsize}}}}
\newcommand{\putdisk}[3]{\put(#1,#2){\circle*{#3}}}
\begin{picture}(180,130)(0,0)
	\putdisk{80}{120}{7}			
	\puttext{70}{120}{{\bf 0}}		
	\puttext{85}{125}{{\rm nw}}		
	\putdisk{20}{80}{7}			
	\puttext{10}{80}{{\bf 1}}		
	\put(20,80){\line(3,2){60}}		
	\puttext{25}{89}{{\rm mo}}		
	\putdisk{60}{80}{7}			
	\puttext{50}{80}{{\bf 3}}		
	\put(60,80){\line(1,2){20}}		
	\puttext{65}{85}{{\rm ll}}		
	\putdisk{100}{80}{7}			
	\puttext{90}{80}{{\bf 6}}		
	\put(100,80){\line(-1,2){20}}		
	\puttext{105}{85}{{\rm lw}}		
	\putdisk{140}{80}{7}			
	\puttext{130}{80}{{\bf 2}}		
	\put(140,80){\line(-3,2){60}}		
	\puttext{145}{85}{{\rm nc}}		
	\putdisk{0}{40}{7}			
	\puttext{-10}{40}{{\bf 4}}		
	\put(0,40){\line(1,2){20}}		
	\put(0,40){\line(3,2){60}}		
	\puttext{5}{50}{{\rm sk}}		
	\puttext{5}{43}{{\rm lb}}		
	\puttext{5}{27}{{\rm Dg}}		
	\putdisk{40}{40}{7}			
	\puttext{30}{40}{{\bf 7}}		
	\put(40,40){\line(-1,2){20}}		
	\put(40,40){\line(3,2){60}}		
	\puttext{45}{35}{{\rm Le}}		
	\puttext{45}{28}{{\rm Br}}		
	\puttext{45}{21}{{\rm Fr}}		
	\putdisk{120}{40}{7}			
	\puttext{110}{40}{{\bf 5}}		
	\put(120,40){\line(-3,2){60}}		
	\put(120,40){\line(1,2){20}}		
	\puttext{125}{35}{{\rm Bn}}		
	\puttext{125}{28}{{\rm Mz}}		
	\putdisk{160}{40}{7}			
	\puttext{150}{40}{{\bf 8}}		
	\put(160,40){\line(-3,2){60}}		
	\put(160,40){\line(-1,2){20}}		
	\puttext{165}{45}{{\rm 1lg}}		
	\puttext{165}{35}{{\rm SW}}		
	\puttext{165}{28}{{\rm Rd}}		
	\putdisk{80}{0}{7}			
	\puttext{70}{-5}{{\bf 9}}		
	\put(80,0){\line(-2,1){80}}		
	\put(80,0){\line(-1,1){40}}		
	\put(80,0){\line(1,1){40}}		
	\put(80,0){\line(2,1){80}}		
	\puttext{83}{10}{{\rm 2lg}}		
\end{picture}
\end{tabular}
\end{tabular}
\end{center}
\caption{{\bf the upper and lower approximations of the Living formal context} \label{Living:approximation}}
\end{table}
\normalsize

In Table~\ref{assignments} we list the two assignment maps:
the upper approximation conceptual assignment
and
the lower approximation conceptual assignment.
These define two conceptual indiscernibility relations
--- the conceptual indiscernibility of possibility defined as the kernel of the upper approximation conceptual assignment,
and the conceptual indiscernibility of necessity defined as the kernel of the lower approximation conceptual assignment.
\scriptsize
\begin{table}
\begin{center}
\begin{tabular}{c@{\hspace{5mm}}c}
	\begin{tabular}[t]{|c@{\hspace{5mm}}c@{\hspace{5mm}}l|} \hline
		\multicolumn{3}{|c|}{{\bf upper approximation assignment}}  \\ \hline\hline
		\multicolumn{3}{|c|}{{\bf indiscernibility of possibility}} \\ \hline\hline
		$\{ \mbox{B}_{0}, \mbox{B}_{5} \}$
					& $\mapsto$ &  $\overline{\mbox{B}}_{0}$ \\
		$\{ \mbox{B}_{1}, \mbox{B}_{2}, \mbox{B}_{6} \}$
					& $\mapsto$ &  $\overline{\mbox{B}}_{1}$ \\
		$\{ \mbox{B}_{7} \}$
					& $\mapsto$ &  $\overline{\mbox{B}}_{2}$ \\
		$\{ \mbox{B}_{3}, \mbox{B}_{4}, \mbox{B}_{8}, \mbox{B}_{9} \}$
					& $\mapsto$ &  $\overline{\mbox{B}}_{3}$ \\
		$\{ \mbox{B}_{10} \}$
					& $\mapsto$ &  $\overline{\mbox{B}}_{4}$ \\
		$\{ \mbox{B}_{11}, \mbox{B}_{15} \}$
					& $\mapsto$ &  $\overline{\mbox{B}}_{5}$ \\
		$\{ \mbox{B}_{12}, \mbox{B}_{13}, \mbox{B}_{16} \}$
					& $\mapsto$ &  $\overline{\mbox{B}}_{6}$ \\
		$\{ \mbox{B}_{14}, \mbox{B}_{17} \}$
					& $\mapsto$ &  $\overline{\mbox{B}}_{7}$ \\
		$\{ \mbox{B}_{18} \}$
					& $\mapsto$ &  $\overline{\mbox{B}}_{8}$ \\ \hline
	\end{tabular}
	&
	\begin{tabular}[t]{|c@{\hspace{5mm}}c@{\hspace{5mm}}l|} \hline
		\multicolumn{3}{|c|}{{\bf lower approximation assignment}} \\ \hline\hline
		\multicolumn{3}{|c|}{{\bf indiscernibility of necessity}}  \\ \hline\hline
		$\{ \mbox{B}_{0} \}$
			& $\mapsto$ & $\mbox{\underline{B}}_{0}$ \\
		$\{ \mbox{B}_{1} \}$
			& $\mapsto$ & $\mbox{\underline{B}}_{1}$ \\
		$\{ \mbox{B}_{3} \}$
			& $\mapsto$ & $\mbox{\underline{B}}_{2}$ \\
		$\{ \mbox{B}_{5} \}$
			& $\mapsto$ & $\mbox{\underline{B}}_{3}$ \\
		$\{ \mbox{B}_{2}, \mbox{B}_{6}, \mbox{B}_{7} \}$
			& $\mapsto$ & $\mbox{\underline{B}}_{4}$ \\
		$\{ \mbox{B}_{8} \}$
			& $\mapsto$ & $\mbox{\underline{B}}_{5}$ \\
		$\{ \mbox{B}_{11} \}$
			& $\mapsto$ & $\mbox{\underline{B}}_{6}$ \\
		$\{ \mbox{B}_{12} \}$
			& $\mapsto$ & $\mbox{\underline{B}}_{7}$ \\
		$\{ \mbox{B}_{4}, \mbox{B}_{14} \}$
			& $\mapsto$ & $\mbox{\underline{B}}_{8}$ \\
		$\{ \mbox{B}_{9}, \mbox{B}_{10}, \mbox{B}_{13}, \mbox{B}_{15}, \mbox{B}_{17}, \mbox{B}_{18}, \mbox{B}_{19} \}$
			& $\mapsto$ & $\mbox{\underline{B}}_{9}$ \\ \hline
	\end{tabular}
\end{tabular}
\end{center}
\caption{{\bf approximation assignments \& conceptual indiscernibility} \label{assignments}}
\end{table}
\normalsize
In Table~\ref{assignments} we use the notation
$\overline{\mbox{B}}_i$ for concepts in the upper approximation concept lattice,
and we use the notation
$\mbox{\underline{B}}_i$ for concepts in the lower approximation concept lattice.
Rough equality is the meet of the possibility and necessity conceptual indiscernibility relations.
The only distinct roughly equal concepts are
\begin{center}
$\begin{array}{ccc}
	\mbox{B}_{2} \equiv \mbox{B}_{6} & \mbox{and} & \mbox{B}_{13} \equiv \mbox{B}_{16} .
 \end{array}$
\end{center}

Consider the concept ``limbed animals'',
which is indexed by $\mbox{B}_{2}$ in the original context,
where it has extent ``Bream'', ``Frog'' and ``Dog''
(a ``Bream'' is a European fresh-water fish related to the Carp),
and intent ``needs water'', ``is motile'' and ``has limbs''.
This concept is indexed by $\overline{\mbox{B}}_{1}$ in the upper approximation context,
where it has extent ``Leech'', ``Bream'', ``Frog'' and ``Dog'',
and intent ``needs water'', ``lives on land'', ``is motile'' and ``has limbs''.
This concept gains the object ``Leech'' in the upper approximation context,
since ``Leech'' and ``Frog'' being indiscernibly equivalent,
the object ``Leech'' possibly ``has limbs'' there.
This concept gains the attribute ``lives on land'' in the upper approximation context,
since the implication
``has limbs'' implies ``lives on land''
holds there.
This concept is indexed by $\mbox{\underline{B}}_{4}$ in the lower approximation context,
where it has extent only the object ``Dog'',
and intent ``needs water'', ``lives on land'', ``is motile'', ``has limbs'' and ``suckles young''.
This concept loses the objects ``Bream'' and ``Frog'' in the lower approximation context,
since ``Leech'' and ``Bream'' being indiscernibly equivalent,
the object ``Bream'' does not necessarily (certainly) ``have limbs''
(same for ``Frog'').
This concept gains the attributes ``lives on land'' and ``suckles young'' in the lower approximation context,
since the implications
``has limbs'' implies ``lives on land''
and
``has limbs'' implies ``suckles young''
hold there.

In the crisp Living context of Table~\ref{Living}
the concept ``limbed animals which live on land'' indexed by $\mbox{B}_{6}$
is more specialized than 
the concept ``limbed animals'' indexed by $\mbox{B}_{2}$,
and is distinguished from $\mbox{B}_{2}$ by the characteristic ``lives on land''.
Since ``Bream'' is a fish and does not live on land,
in the lower approximation context
where ``Bream'' and ``Frog'' are indiscernible,
it is not necessary (certain) that a ``Frog'' is a land dwelling organism.
In this context ``lives on land'' is implied by ``has limbs'',
and is no longer a distinguishing characteristic.
Although ``lives on land'' serves as a distinguishing attribute
for concepts $\mbox{B}_{2}$ and $\mbox{B}_{6}$
in the crisp Living context of Table~\ref{Living},
it no longer does in the rough setting of indiscernibility relation~\ref{indiscernibility}.
By the same token,
although the ``lives on land'' attribute
distinguishes the object ``Frog'' from the objects ``Leech'' and ``Bream''
in the crisp Living context of Table~\ref{Living},
it cannot in the rough setting of indiscernibility relation~\ref{indiscernibility}
where these three objects are indiscernible.

\section*{Summary}

This paper has introduced the new theory of Rough Concept Analysis,
which is a synthesis of Rough Sets and Formal Concept Analysis.
Rough Concept Analysis,
which studies the rough approximation of conceptual structures,
provides an ``approximation theory'' for knowledge representation and knowledge discovery.
The notions of upper and lower approximations were extended
from subsets of objects to formal contexts,
which are viewed here as attribute-indexed collections of subsets of objects.
Upper and lower formal approximation contexts
were used to provide external notions of upper and lower approximation
for formal concepts.
Since these conceptual approximations were shown to be
join and meet-preserving monotonic functions between concept lattices,
a notion of rough conceptual join can be defined via upper approximation
and
a notion of rough conceptual meet can be defined via lower approximation.
All of these notions were illustrated 
by a simple example concerning the ecology of Living organisms.

Data modeling with distributed constraints extends 
the notion of conceptual scaling in Formal Concept Analysis 
by combining it with notions from Entity-Relationship database modeling
\cite{kent93b,kent94c}.
Formal contexts have been shown to be special cases of distributed constraints
(namely, single-sorted distributed constraints),
whereas
distributed constraints are interpretable via the notion of satisfaction 
in terms of formal contexts.
In this model,
formal concepts correspond to database relations,
and conceptual meet corresponds to natural join.
In future work
I will give a version of contextual approximation
in terms of contextual flow along description functions.
Using this approach 
I will extend upper and lower approximation
from formal contexts to distributed constraints.
Then,
rough formal concepts will correspond to rough database relations,
and rough conceptual meet will correspond to rough natural join.



\begin{thebibliography}{1}

\bibitem{kent93b}
R.\ Kent and J.\ Brady, 
``Formal Concept Analysis with Many-Sorted Attributes'',
{\sl Proceedings of the 5th International Conference on Computing and Information\/},
Sudbury, Ontario, Canada, 1993.

\bibitem{kent94c}
R.\ Kent and F.\ Vogt and R.\ Wille, 
``Data Modeling with Constraints'' (unpublished),
	This paper is based upon a series of lectures given by the first author
	to the Research Group in Formal Concept Analysis at the Technische Hochschule in Darmstadt, Germany,
	in the summer of 1993,
	when the first author was on a guest researchership there.

\bibitem{lambek58} 
J.\ Lambbek,
``The Mathematics of Sentence Structure'',
{\sl American Mathematical Monthly\/} 1958,
vol.\ 65.

\bibitem{pawlak82}
Z.\ Pawlak,
``Rough Sets'',
{\sl International Journal of Information and Computer Science\/} 1982,
vol.\ 11, pp.\ 341--356.

\bibitem{wille82} 
R.\ Wille,
``Restructuring Lattice Theory: An Approach Based on Hierarchies of Concepts'',
{\sl Ordered Sets\/},
I.\ Rival ed.,
pp.\ 445--470,
Reidel,
Dordrecht-Boston,
1982.

\end{thebibliography}
\end{document}